\documentclass{article}
\pdfoutput=1

\PassOptionsToPackage{numbers, compress}{natbib}
\usepackage[preprint]{neurips_2023}

\usepackage[utf8]{inputenc} 
\usepackage[T1]{fontenc}    
\usepackage{hyperref}       
\usepackage{url}            
\usepackage{booktabs}       
\usepackage{amsfonts}       
\usepackage{nicefrac}       
\usepackage{microtype}      
\usepackage{xcolor}         

\usepackage{pythonhighlight}
\usepackage{graphicx}
\usepackage{multirow}
\usepackage{subfigure}
\usepackage{bbding}

\title{Student-friendly Knowledge Distillation}

\author{
Mengyang Yuan \quad Bo Lang \quad Fengnan Quan~~~
\smallskip 
\\
School of Computer Science and Engineering, Beihang University
\\
Beijing, China
\\
  \texttt{\{yuanmengyang, langbo, quanfengnan\}@buaa.edu.cn}
}

\begin{document}

\maketitle

\begin{abstract}
In knowledge distillation, the knowledge from the teacher model is often too complex for the student model to thoroughly process. However, good teachers in real life always simplify complex material before teaching it to students. Inspired by this fact, we propose student-friendly knowledge distillation (SKD) to simplify teacher output into new knowledge representations, which makes the learning of the student model easier and more effective. SKD contains a softening processing and a learning simplifier. First, the softening processing uses the temperature hyperparameter to soften the output logits of the teacher model, which simplifies the output to some extent and makes it easier for the learning simplifier to process. The learning simplifier utilizes the attention mechanism to further simplify the knowledge of the teacher model and is jointly trained with the student model using the distillation loss, which means that the process of simplification is correlated with the training objective of the student model and ensures that the simplified new teacher knowledge representation is more suitable for the specific student model. Furthermore, since SKD does not change the form of the distillation loss, it can be easily combined with other distillation methods that are based on the logits or features of intermediate layers to enhance its effectiveness. Therefore, SKD has wide applicability. The experimental results on the CIFAR-100 and ImageNet datasets show that our method achieves state-of-the-art performance while maintaining high training efficiency.
\end{abstract}

\section{Introduction}
In recent years, deep neural networks have achieved great success in many computer vision tasks, such as image classification \cite{res, mo, sqex, shu2, verydeep}, object recognition \cite{maskrcnn, fasterrcnn, fpn}, and semantic segmentation \cite{atrous, fcn, pspn}. As the performance of neural network models improves, their computational and storage costs also increase, making model compression an important research problem \cite{compression}. Knowledge distillation is an important model compression method \cite{kd}.

\begin{figure}
\subfigure[Vanilla KD\label{fig:kdil}]
{
    \begin{minipage}{.37\linewidth}
        \centering
        \includegraphics[scale=0.54]{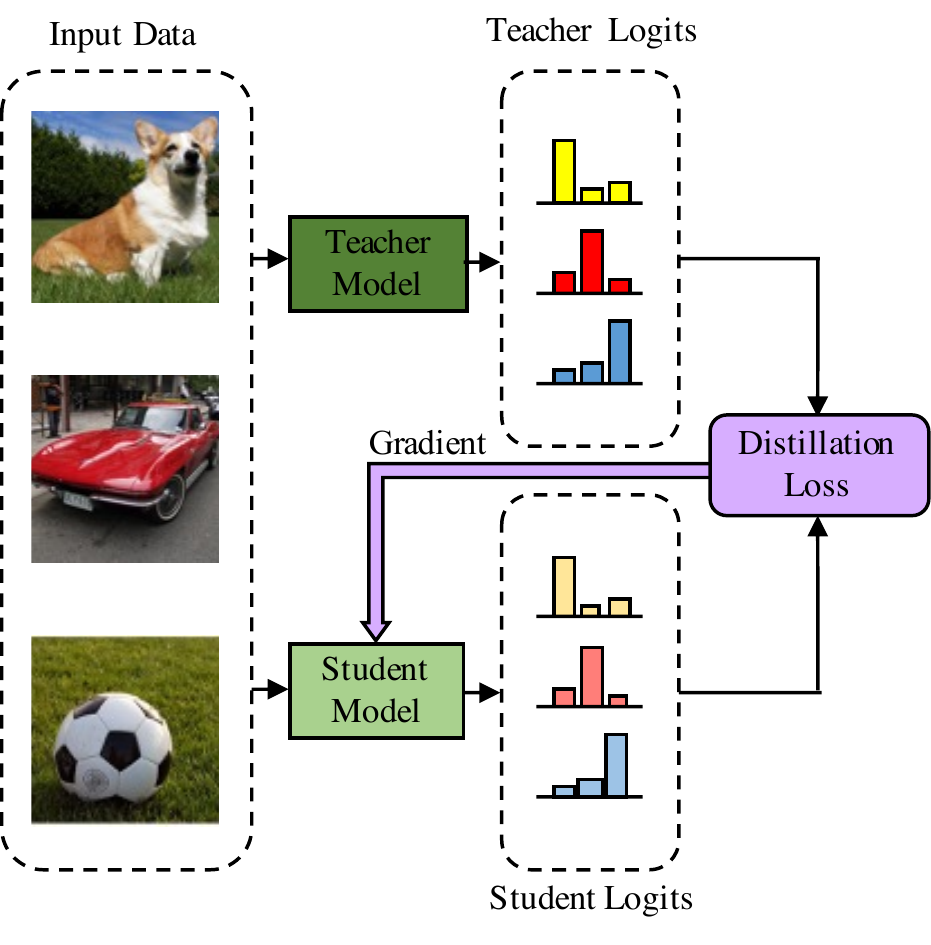}
    \end{minipage}
}
\subfigure[Student-friendly Knowledge Distillation (SKD)\label{fig:skdil}]
{
 	\begin{minipage}{.6\linewidth}
        \centering
        \includegraphics[scale=0.54]{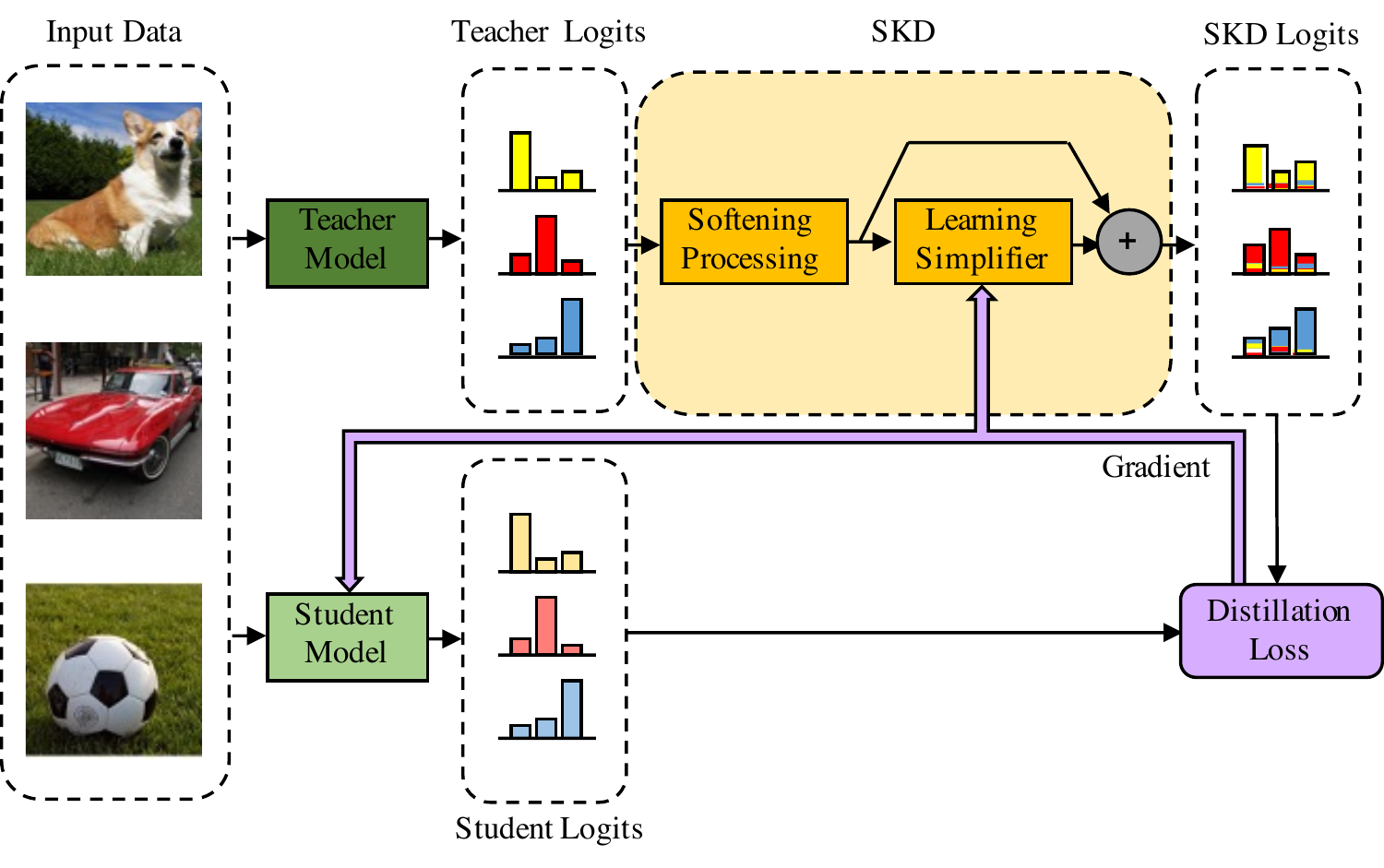}
    \end{minipage}
}
	\caption{Illustration of vanilla knowledge distillation (KD) and our student-friendly knowledge distillation (SKD). (a) KD uses the outputs of the teacher and student models to calculate the distillation loss. (b) Our SKD transforms the outputs of the teacher model to obtain the new SKD logits, which are then compared with the logits of the student model to calculate the distillation loss. The gradient obtained through backpropagation optimizes the student model and simultaneously optimizes the learning simplifier. Our SKD achieves better results than KD using the same loss function.}
\end{figure}

Knowledge distillation enables a smaller model with fewer parameters (the student model) to learn from a larger model (the teacher model) to achieve better performance. The vanilla knowledge distillation (KD) method uses the Kullback-Leibler (KL) divergence \cite{kl} to mimic the teacher model’s logits using the student model \cite{kd}, as shown in Figure \ref{fig:kdil}. The student model learns from the logits of the teacher model to improve its performance. As researchers increasingly studied knowledge distillation, they enabled the student model to learn from the features of intermediate layers of the teacher model \cite{ofd, crd, reviewkd, fitnet, rkd}. However, as the outputs of intermediate layers differ across deep learning models, the design complexity and computational costs of feature-based methods increase. Recently, some researchers have begun investigating distillation methods based on the logits of the teacher model \cite{dkd, dist, nkd}. The distillation loss is modified to enable the student model to effectively utilize knowledge of the teacher model, achieving distillation results comparable to or even superior to those of feature-based methods.

In logit-based methods, the logits of both the teacher and student models are softened using temperature, which leads to a softer label distribution, reduces the gap between the target class and other classes, and allows the distillation loss to focus more on other classes, thereby improving the training effect of the student model \cite{kd}. However, even with temperature, student models still cannot closely imitate a teacher model’s logits due to insufficient capacity and limited data \cite{dkdrw}.

In real life, good teachers often simplify new knowledge according to students' abilities to help them better understand it. Based on the educational experience of human teachers, we propose a new method called student-friendly knowledge distillation (SKD), outlined in Figure \ref{fig:skdil}, to optimize the output knowledge of the teacher model, making it easier for students to learn.

SKD utilizes the learning simplifier to transform the output distribution of the teacher model into a new distribution that serves as the learning target for the student network. During the training process, the learning simplifier and the student model are jointly optimized using the distillation loss for gradient backpropagation. This allows the new logit distribution to better fit the characteristics of the student model, making it easier for the student model to imitate the teacher model. We design the learning simplifier using self-attention to better construct a simplified logit distribution for the student. The self-attention mechanism enables SKD to adjust the logit distribution of the teacher model along with the output of the student model by using the similarity relationships among the data in the output of the teacher model. This makes it easier for the student network to imitate the simplified logit distribution and learn the knowledge of the teacher network. To improve the learning effect of the learning simplifier on the relationships between data, we incorporate softening processing at the beginning of SKD. We use the temperature-scaled $\mathrm{LogSoftmax}$ function to soften the output of the teacher model, similar to temperature softening in the distillation loss. Larger models tend to produce sharper output distributions than smaller models \cite{confcal}, which means that the softened label distribution is more suitable for smaller student models to learn. We conducted extensive experiments, and the results showed that our SKD achieved the best performance in many combinations of knowledge distillation models.

Furthermore, most existing knowledge distillation methods aim to improve either the distillation of the intermediate features or the distillation loss function of the logits. However, our SKD changes the logits of the teacher model without changing the distillation loss function. Therefore, we can use SKD in conjunction with existing knowledge distillation methods. Experimental results show that the combined method significantly improves upon the original methods, resulting in better-performing student models.

To summarize, the main contributions of our paper are as follows:
\begin{itemize}
	\item We re-evaluate the knowledge representation of the teacher model in logit-based knowledge distillation methods. Inspired by real-life teaching scenarios, to enhance the learning performance of the student model, we propose a new direction for improving knowledge distillation. The core of this new direction is to consider the poor capacity of the student model and decrease the knowledge difficulty of the teacher model accordingly.
	\item Based on the above principle, we propose a knowledge distillation method called SKD. SKD uses temperature softening to enable the teacher model to utilize a higher distillation temperature than the student model, resulting in smoother teacher output. We also create a learning simplifier to further simplify the teacher output based on the similarity relationships among the data. By jointly optimizing the learning simplifier and the student model using distillation loss, the teacher's output can better adapt to the student model, reducing the student's learning difficulty. The experimental results demonstrate that SKD achieves state-of-the-art performance while maintaining high training efficiency.
	\item Our proposed SKD uses the same distillation loss as KD for optimization, which makes it easy to blend with other knowledge distillation models, including feature-based or logit-based distillation methods. The experimental results show that SKD can improve the performance of student networks in present methods and achieve the current best results.
\end{itemize}

\section{Preliminaries}
\subsection{Vanilla knowledge distillation}
The process of vanilla knowledge distillation (KD) \cite{kd} is shown in Figure \ref{fig:kdil}. For the training data $x$ with the label $y$ in a dataset with $K$ classes, the outputs of the teacher model and the student model are $g^t\in\mathbb{R}^K$ and $g^s\in\mathbb{R}^K$, respectively. Using the softmax function yields the student's prediction $p^s=\mathrm{softmax}\left(g^s\right)\in\mathbb{R}^K$, and we can then compute the cross-entropy loss between the student's prediction and the ground-truth label:
\begin{equation} \label{eq:lce}
	\mathcal{L}_{\mathrm{CE}}=\sum_{i=1}^K y_i \log(p_i^s)\label{eq:KL}.
\end{equation}
Using the softmax function with temperature, we obtain the softened teacher prediction ${\widetilde{p}}^t=\mathrm{softmax}\left(g^t/T\right)\in\mathbb{R}^K$ and the softened student prediction ${\widetilde{p}}^s=\mathrm{softmax}\left(g^s/T\right)\in\mathbb{R}^K$. Through the temperature, the predictions become smoother over each class, so the distillation loss is better able to reflect the differences between the other classes in addition to the correct one. Then, we can compute the distillation loss between the softened predictions with the KL divergence:
\begin{equation} \label{eq:kl}
	\mathcal{L}_{\mathrm{KL}}=\mathrm{KL}(\widetilde{p}^t||\widetilde{p}^s)=\sum_{i=1}^K \widetilde{p}^t_i \log(\frac{\widetilde{p}^t_i}{\widetilde{p}^s_i} ).
\end{equation}
The total loss of KD is:
\begin{equation}
	\mathcal{L}_{total}=\alpha \mathcal{L}_{\mathrm{CE}}+\beta \mathcal{L}_{\mathrm{KL}},
\end{equation}
where $\alpha$ and $\beta$ are coefficients used to balance the two parts. Knowledge distillation optimizes the student model by optimizing this loss function.
\subsection{Attention}
The attention used in our SKD is the standard self-attention in the Transformer \cite{trans}. First, the input data are encoded through a linear projection $\mathrm{Linear}_1$ to obtain the corresponding query $Q$, key $K$, and value $V$ of dimension $D$. Then, based on the query and key, the attention matrix $A$ relating them is calculated:
\begin{equation}
	A=\mathrm{softmax}(QK^\top/\sqrt{D} ).
\end{equation}
Then, the weighted sum of values is calculated based on the attention matrix $A$, and encoded through another linear projection $\mathrm{Linear}_2$ to obtain the output of the self-attention:
\begin{equation}
	Output=\mathrm{Linear}_2(AV).
\end{equation}
Through self-attention, new representations of the data based on the relationships among these data are obtained.

\section{Methods}
\label{Methods}
\subsection{Motivation}
In knowledge distillation, the capacity of the teacher model is greater than that of the student model, making it difficult for the student model to accurately simulate the output distribution of the teacher model. In real life, good teachers simplify complex knowledge before teaching it to their students. Inspired by this fact, we propose the student-friendly knowledge distillation (SKD) method, as shown in Figure \ref{fig:skdil}. As it is difficult for the student model to generate complex and sharp outputs such as those of the teacher model \cite{confcal}, our SKD first softens the output of the teacher model via softening processing, making it easier for the student model to learn and more advantageous for the learning simplifier to handle.

The learning simplifier in SKD is used to modify the teacher output to reduce the difficulty of the student model to mimic the teacher model's output. The learning simplifier and the student model jointly optimize the distillation loss. The learning simplifier uses the real-time logits of the student model as its optimization objective. Therefore, the learning simplifier can transform the output of the teacher model, which is difficult for the student model to mimic, into a distribution that is more similar to the student model's output, thus reducing the difficulty of the student model to mimic the teacher model's output. By using SKD to make minor changes to the output of the teacher model, the student model can better mimic the teacher model, thereby improving the effectiveness of knowledge distillation.

\subsection{Overall}
Our SKD performs softening processing and the learning simplifier on the output of the teacher model to obtain a new teacher distribution $g^{\mathrm{SKD}}$. The calculation process is shown in Figure \ref{fig:skdil}.

The output of the teacher model $g^t\in\mathbb{R}^{K}$ is first softened to obtain the softened logit distribution:
\begin{equation}
	g^t_{soft}=\mathrm{Softening}(g^t).
\end{equation}
Then, through the learning simplifier, the change in logits is obtained:
\begin{equation}
	\Delta_{Simplifier}=\mathrm{Simplifier}(g^t_{soft}).
\end{equation}
$\Delta_{Simplifier}$ is added to the softened teacher logits $g_{soft}^t$, and the output distribution of SKD is obtained:
\begin{equation}
	g^{\mathrm{SKD}}=\Delta_{Simplifier}+g_{soft}^t.
\end{equation}
Using the softmax function with temperature, we obtain the softened prediction of SKD $\widetilde{p}^{\mathrm{SKD}}=\mathrm{softmax}\left(g^{\mathrm{SKD}}/T\right)\in\mathbb{R}^K$. Finally, similar to the original knowledge distillation method, the distillation loss $\mathcal{L}_{\mathrm{SKD}}$ can be calculated using the softmax function with temperature and the KL divergence, as in Eq. \ref{eq:kl}:
\begin{equation}\label{eq:lskd}
	\mathcal{L}_{\mathrm{SKD}}=\mathrm{KL}(\widetilde{p}^{\mathrm{SKD}}||\widetilde{p}^{s}).
\end{equation}
Therefore, the total loss of SKD is:
\begin{equation}
	\mathcal{L}_{total}=\mathcal{L}_{\mathrm{CE}}+\alpha \mathcal{L}_{\mathrm{SKD}},
\end{equation}
where $\mathcal{L}_{\mathrm{CE}}$ is the cross-entropy loss between the student model's predictions and the true labels, as defined in Eq. (\ref{eq:lce}). To facilitate parameter tuning, the coefficient of the cross-entropy loss is fixed at 1.0. $\alpha$ is the weight coefficient of the SKD loss, which adjusts the relative contributions of the distillation loss and the cross-entropy loss. The effect of the distillation loss is to make the student model mimic the output of the teacher model. The larger the value of $\alpha$, the more the student model needs to pursue the output of the teacher model during training. However, if the gap between the teacher model and the student model is large, the student model will find it difficult to mimic the output of the teacher model. For models of the same type, the smaller the difference in performance between the teacher and student models, the larger the optimal value of $\alpha$, which enables the student model to mimic the output of the teacher model better. However, the optimal value of $\alpha$ still needs to be determined through experiments.
%
%
\subsection{Learning simplifier}
For the design of the learning simplifier, we considered using either fully connected (FC) layers or self-attention, both of which can construct output distributions that are closer to the student model based on the output of the teacher model. The student model mainly learns the relationships between categories in logit-based knowledge distillation methods. Unlike FC layers, which only consider the input of a single data point, self-attention can learn the relationships between each batch of data and weight their values according to the relationships to obtain the final output. We conducted experiments comparing the different implementation methods on the implementation methods on the CIFAR-100 dataset, where the teacher model is ResNet32$\times$4 and the student model is ResNet8$\times$4, and the results are shown in Table \ref{tab:ls}. Based on the results, we choose self-attention to implement our learning simplifier.
\begin{table}
	\begin{minipage}[t]{0.5\textwidth}
\centering
\caption{Comparison of different implementations of the learning simplifier.\label{tab:ls}}
  \begin{tabular}{cc}
    Learning Simplifier     & Top-1 $(\%)$ \\
    \midrule[1pt]
    1-layer FC & 75.18     \\
    2-layer FC     & 75.28      \\
     \textbf{Self-attention}     &  \textbf{75.83}  \\
  \end{tabular}
\end{minipage}
\begin{minipage}[t]{0.5\textwidth}
\centering
\caption{Softening Processing Effectiveness.\label{tab:soft}}
  \begin{tabular}{ccc}
    Softening     &Temperature& Top-1 $(\%)$ \\
    \midrule[1pt]
     \XSolidBrush & - &75.83     \\
    \Checkmark     &3.0&  76.42     \\
   	\Checkmark     &4.0&  \textbf{76.84}     \\
   	\Checkmark     &5.0&  76.51     \\
  \end{tabular}
\end{minipage}
\end{table}
\subsection{Softening processing}
The self-attention used in our learning simplifier focuses on the relationships among input data. Teacher models often have high confidence, resulting in the output distribution being dominated by the target class, with little difference in output distribution between similar classes. This makes it difficult for self-attention to learn the relationships among different data of the same class. In knowledge distillation, the logits are softened by temperature, which smooths the label distribution so that the logits can reflect the relationships among classes other than the target class \cite{kd}. Therefore, inputting the distribution softened by temperature into self-attention can improve the learning simplifier’s ability to learn the relationships among different data, thereby enhancing the effectiveness of SKD.

On the other hand, using temperature to soften the logits of the teacher model is equivalent to using a higher temperature for the teacher model than for the student model, rather than using the same distillation temperature for both as in vanilla knowledge distillation; consequently, the output distribution of the teacher model is softer. Because models with more parameters tend to have sharper output distributions after training than models with fewer parameters \cite{confcal}, smaller student models more easily imitate softened distributions.

We conducted experiments to verify the effectiveness of using a temperature-scaled $\mathrm{LogSoftmax}$ function on the CIFAR-100 dataset, where the teacher model is ResNet32$\times$4 and the student model is ResNet8$\times$4. Table \ref{tab:soft} shows the results. Softening the output of the teacher model with a $\mathrm{LogSoftmax}$ function with a temperature set to $4.0$ leads to better results with our SKD.

\subsection{Combination with other methods}
Notably, our SKD improves knowledge distillation by modifying the logits of the teacher model. As shown in \ref{eq:lskd}, SKD uses the same distillation loss as vanilla KD. Therefore, SKD can be easily combined with other logit-based methods that modify the distillation loss. The modified logits in SKD can be used as the logits of the teacher model in other methods. Fusing SKD with other models can improve their effectiveness.

On the other hand, SKD only modifies the logits of the teacher model and does not change the structure of the intermediate layers in the model. Therefore, it does not conflict with feature-based methods. By adding the distillation losses, SKD is easily combined with feature-based methods to improve the performance of the original methods.

According to our experiments in section \ref{sec:combin}, integrating SKD with the currently best-performing distillation methods from two categories results in state-of-the-art performance.

\section{Experiments}
We conducted comprehensive experiments on image classification tasks on \textbf{CIFAR-100} \cite{cifar} and \textbf{ImageNet} \cite{imagenet}. The detailed experimental settings are given in Appendix \ref{app:seting}.

\subsection{Comparison with state-of-the-art methods}
The results on the \textbf{CIFAR-100} dataset are shown in Table \ref{tab:cifar_same} and Table \ref{tab:cifar_dif}, where Table \ref{tab:cifar_same} shows the results where the teacher and student models had the same architectures, and Table \ref{tab:cifar_dif} shows the results where the teacher and student models had different architectures.
\begin{table}
  \centering
  \caption{Results on the \textbf{CIFAR-100} dataset. Teacher and student models have the \textbf{same} architectures. All reported accuracy results are averaged over five trials. $\Delta$ denotes the improvement of SKD over the KD method. The results marked in red and blue are the best and second-best results, respectively.\label{tab:cifar_same}}
    \resizebox{1\columnwidth}{!}{
\begin{tabular}{ccccccc}
\multirow{2}*{Teacher}	& ResNet32$\times$4	& ResNet110 & ResNet56 	& WRN-40-2 	& WRN-40-2 & VGG13 \\
& 79.55 	& 74.31 	& 72.34 	& 75.61 	& 75.61 & 74.64 \\
\multirow{2}*{Student} & ResNet8$\times$4 & ResNet32 & ResNet20 & WRN-16-2 & WRN-40-1 & VGG8 \\ 
& 72.50 & 71.14 & 69.06 & 73.26 & 71.98 & 70.36 \\
\midrule[1pt]
\multicolumn{7}{c}{Feature-based methods}\\
FitNet \cite{fitnet} & 73.52 & 70.98 & 69.02 & 73.59 & 72.08 & 71.37 \\
RKD \cite{rkd} & 72.50 & 71.90 & 69.81 & 72.91 & 71.80 & 70.43 \\ 
CRD \cite{crd} & 75.73 & 73.71 & 71.31 & 75.66 & 74.36 & 73.90\\ 
OFD \cite{ofd} & 74.88 & 72.78 & 69.96 & 75.50 & \textbf{\textcolor{red}{75.23}} & 73.30 \\ 
ReviewKD \cite{reviewkd} & 75.68 & \textbf{\textcolor{blue}{73.73}} & 71.23 & \textbf{\textcolor{blue}{76.28}} & \textbf{\textcolor{blue}{75.11}} & 73.80 \\ 
\multicolumn{7}{c}{Logits-based methods}\\
KD \cite{kd} & 73.56 & 73.42 & 71.08 & 75.01 & 73.75 & 73.43 \\ 
DKD \cite{dkd} &  \textbf{\textcolor{blue}{76.13}} & 73.71 & \textbf{\textcolor{blue}{71.64}} & 75.52 & 74.43 & \textbf{\textcolor{blue}{74.57}} \\ 
 \textbf{SKD} &  \textbf{\textcolor{red}{76.84}} &  \textbf{\textcolor{red}{74.06}} &  \textbf{\textcolor{red}{71.73}} &  \textbf{\textcolor{red}{76.29}} &  74.52 &  \textbf{\textcolor{red}{74.94}} \\ 
$\Delta$ & \textcolor{green}{+3.28} & \textcolor{green}{+0.64} & \textcolor{green}{+0.65} & \textcolor{green}{+1.28} & \textcolor{green}{+0.77} & \textcolor{green}{+1.51} \\ 
\end{tabular}
}
\end{table}

\begin{table}
    \centering
    \caption{Results on the \textbf{CIFAR-100} dataset. Teacher and student models have \textbf{different} architectures. All reported accuracy results are averaged over five trials. $\Delta$ denotes the improvement of SKD over the KD method. The results marked in red and blue are the best and second-best results, respectively.\label{tab:cifar_dif}}
    \resizebox{1\columnwidth}{!}{
    \begin{tabular}{cccccc}
        \multirow{2}*{Teacher} & ResNet32$\times$4 & WRN-40-2 & ResNet50 & VGG13 & ResNet32$\times$4 \\ 
        & 79.55 & 75.61 & 79.34 & 74.64 & 79.55 \\
        \multirow{2}*{Student} & ShuffleNet-V2 & ShuffleNet-V1 & MobileNet-V2 & MobileNet-V2 & VGG8 \\ 
        & 71.82 & 70.50 & 64.60 & 64.60 & 70.36 \\
        \midrule[1pt]
		\multicolumn{6}{c}{Feature-based methods}\\

        FitNet \cite{fitnet} & 74.29 & 73.54 & 63.11 & 63.66 & 71.72 \\ 
        RKD \cite{rkd} & 74.08 & 73.27 & 65.05 & 64.90 & 70.90 \\ 
        CRD \cite{crd} & 76.04 & 75.94 & 69.55 & \textbf{\textcolor{blue}{69.36}} & 73.65 \\ 
        OFD \cite{ofd} & 77.09 & 76.63 & 65.81 & 65.23 & 73.52 \\ 
        ReviewKD \cite{reviewkd} & \textbf{\textcolor{red}{77.19}} & \textbf{\textcolor{red}{77.40}} & 67.07 & 69.00 & 74.19 \\
        \multicolumn{6}{c}{Logits-based methods}\\
        KD \cite{kd} & 75.37 & 75.52 & 68.73 & 68.02 & 72.70 \\ 
        DKD \cite{dkd} & 76.90 & 76.65 & \textbf{\textcolor{red}{70.46}} & \textbf{\textcolor{red}{69.41}} & \textbf{\textcolor{blue}{74.32}} \\ 
        \textbf{SKD} & \textbf{\textcolor{blue}{76.96}} & \textbf{\textcolor{blue}{76.96}} & \textbf{\textcolor{blue}{69.99}} & 69.25 & \textbf{\textcolor{red}{74.59}} \\
        $\Delta$ & \textcolor{green}{+1.59}	 & \textcolor{green}{+1.44}	 & \textcolor{green}{+1.26}	 & \textcolor{green}{+1.23}	 & \textcolor{green}{+1.89} \\
    \end{tabular}
    }
\end{table}

Comparing the experimental results of KD with those of SKD, SKD performs significantly better when using the same distillation loss function as KD. The largest improvement occurred when the teacher model was ResNet32$\times$4 and the student model was ResNet8$\times$4. The improvement reached 3.28$\%$. This indicates that SKD can improve the learning performance of the student model simply by modifying the logits of the teacher model.

Of the six experiments where the teacher and student models had the same architectures, SKD achieved the best performance compared to all feature-based and logit-based methods in five experiments. It ranked third in only one experiment where the feature-based method OFD \cite{ofd} and ReviewKD \cite{reviewkd} performed better. Of the five experiments where the teacher and student models were of different types, SKD achieved the best performance in one experiment, ranked second in three experiments, and ranked third in one experiment.

Furthermore, SKD outperformed other state-of-the-art knowledge distillation methods, such as ReviewKD \cite{reviewkd} and DKD \cite{dkd}, in 8 and 9 out of 11 teacher-student model combination experiments, respectively, achieving the best performance. Only in one teacher-student model combination did SKD not achieve a top-two result. This suggests that SKD can achieve the best distillation effect while keeping the design and training simple.

The results on the \textbf{ImageNet} dataset are shown in Table \ref{tab:imagenet}. Our SKD continued to perform significantly better than the classical KD. Moreover, compared with other distillation methods, SKD achieved the first and second-best results in two experiments based on top-1 accuracy and the second and third-best results in two experiments based on top-5 accuracy. This suggests that the performance of SKD is superior to that of most of the current best methods and that it achieved the best performance among logit-based methods.

\begin{table}
    \centering
    \caption{Results on the \textbf{ImageNet} dataset. The SKD results are averaged over three trials. The results of other methods are cited in \cite{dkd}. $\Delta$ denotes the improvement of SKD over the KD method. The results marked in red and blue are the best and second-best results, respectively.\label{tab:imagenet}}
    \resizebox{1\columnwidth}{!}{
    \begin{tabular}{cc|cccc|cccc}
    	& &\multicolumn{4}{c}{Feature-based methods}&\multicolumn{3}{|c}{Logits-based methods}\\
    	 \multicolumn{2}{c|}{Teacher(Student)} & AT \cite{at} & OFD \cite{ofd} & CRD \cite{crd} & ReviewKD \cite{reviewkd} & KD \cite{kd} & DKD \cite{dkd} & \textbf{SKD} & $\Delta$\\
    	 \midrule[1pt]
    	ResNet34&Top-1& 70.96 & 70.81 & 71.17 & 71.61 & 70.66 & \textbf{\textcolor{blue}{71.70}} & \textbf{\textcolor{red}{71.86}} & \textcolor{green}{+1.20} \\
    	(ResNet18)&Top-5& 90.01 &89.98 & 90.13 & \textbf{\textcolor{red}{90.51}} & 89.88 & 90.41 & \textbf{\textcolor{blue}{90.44}} & \textcolor{green}{+0.56}\\
    	\midrule
    	ResNet50&Top-1& 69.56 & 71.25 & 71.37 & \textbf{\textcolor{red}{72.56}} & 68.58 & 72.05  & \textbf{\textcolor{blue}{72.24}} & \textcolor{green}{+3.66}\\
    	(MobileNet-V2)&Top-5& 89.33 & 90.34 & 90.41 & \textbf{\textcolor{blue}{91.00}} & 88.98 & \textbf{\textcolor{red}{91.05}}  & 90.56 & \textcolor{green}{+1.58}\\
    \end{tabular}
    }
\end{table}
\subsection{Combination with other methods}
\label{sec:combin}
We combined SKD with the current best-performing logit-based method DKD \cite{dkd} and the feature-based method ReviewKD \cite{reviewkd} and performed experiments on the CIFAR-100 dataset. The experimental results are shown in Table \ref{tab:combin}. Using SKD in combination with the two other methods significantly improves the student model's accuracy, and the combined model achieved better performance than the two standalone methods. This strongly verifies the effectiveness of SKD and its compatibility with other knowledge distillation methods.
\begin{table}
    \centering
    \caption{Accuracy ($\%$) of SKD combined with other methods. All reported accuracy results are averaged over five trials on the CIFAR-100 dataset. $\Delta$ represents the difference in the accuracy before and after fusion with SKD.\label{tab:combin}}
    \begin{tabular}{ccccc}
        \multirow{2}*{Teacher} & ResNet32$\times$4& VGG13 & ResNet32$\times$4 & VGG13 \\ 
         & 79.55 & 74.64 & 79.55 & 74.64 \\ 
        \multirow{2}*{Student} & ResNet8$\times$4 & VGG8 & 	ShuffleNet-V2 & MobileNet-V2 \\ 
         & 72.50 & 70.36 & 71.82 & 64.60 \\ 
        \midrule[1pt]
        ReviewKD \cite{reviewkd} & 75.68 & 73.80 & 77.19 & 69.00 \\
        \textbf{SKD+ReviewKD	}& \textbf{77.20	}& \textbf{75.07} &\textbf{77.35}	&\textbf{69.82}\\
        $\Delta$ & \textcolor{green}{+1.52} & \textcolor{green}{+1.27} & \textcolor{green}{+0.16} & \textcolor{green}{+2.75}\\
        \midrule
        DKD \cite{dkd} & 76.13 & 74.57 & 76.90 & 69.41 \\ 
        \textbf{SKD+DKD} & \textbf{76.68} & \textbf{75.15} & \textbf{77.52} & \textbf{69.44} \\ 
        $\Delta$ & \textcolor{green}{+0.55} & \textcolor{green}{+0.58} & \textcolor{green}{+0.62} & \textcolor{green}{+0.03} \\ 
    \end{tabular}
\end{table}

\section{Analysis}
To elucidate the principles of SKD, we conducted analyses of five aspects of SKD: (1) the changes in the logits, (2) the distillation fidelity of the student model, (3) the visualization of the student model's output features, (4) the attention matrix in the learning simplifier (see Appendix \ref{app:ana}), and (5) the training efficiency of SKD (see Appendix \ref{app:ana}). The dataset used in the experiment of this section was CIFAR-100, the teacher model was ResNet32$\times$4, and the student model was ResNet8$\times$4.
\paragraph{Changes in the logits}
By observing the changes in the logits before and after the application of SKD, we can see the changes in the learning objectives of the student model. Based on the attention matrix, we obtained the final output of the learning simplifier, which is the change in the logits. We calculated the average change in the target class and other classes of the data distribution on the training set, and the results are shown in Table \ref{tab:lsout}. Compared with the value for other classes, the target class value is significantly reduced by the learning simplifier. This allows the student model to learn the relationships between other classes more effectively when using distillation loss for training, thereby improving the learning effectiveness of the student model.
\begin{table}[htbp]
\begin{minipage}[t]{0.48\linewidth}
	\centering
	\caption{The outputs of the learning simplifier for different classes.\label{tab:lsout}}
  \begin{tabular}{cc}
    Class     & $\Delta$ \\
    \midrule[1pt]
      Target & -0.38     \\
   	Others & +0.02      \\
  \end{tabular}
\end{minipage}
\hspace{0.01\linewidth}%
\begin{minipage}[t]{0.48\linewidth}
	\centering
	\caption{Accuracy before and after the application of SKD.\label{tab:skdac}}
  	\begin{tabular}{ccc}
    SKD     & Top-1 $(\%)$& Top-5 $(\%)$\\
    \midrule[1pt]
      \XSolidBrush & 79.55 & 94.62  \\
   	\Checkmark     & 79.55 & 94.62      \\
  \end{tabular}
\end{minipage}
\end{table}

We also conducted experiments on the accuracy of the output of the teacher model before and after SKD was applied on the validation set, as shown in Table \ref{tab:skdac}. We found that SKD did not change the accuracy of the teacher model, indicating that SKD did not improve the accuracy of the teacher's knowledge. Instead, SKD reduced the learning difficulty for the student model based on the relationships between classes in the output of the teacher model. This led to an improvement in the performance of knowledge distillation.

\begin{figure}[htbp]
    \centering
    \includegraphics[width=1\linewidth]{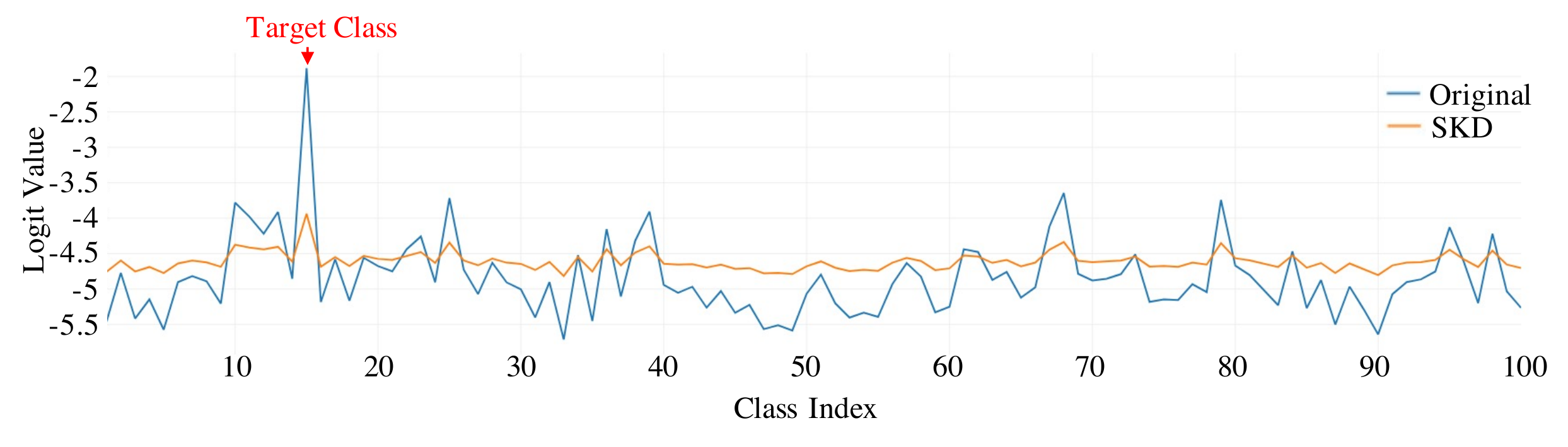}
  	\caption{Visualization of the teacher logits and the SKD logits for 100 classes on a random image.\label{fig:logits}}
\end{figure}

To study the changes in the output of the teacher model caused by SKD, we visualized the original teacher logit distribution and the SKD logit distribution. To make the visualization results clearer, we visualized the logits processed by the temperature-scaled $\mathrm{LogSoftmax}$ function. The visualization results are shown in Figure \ref{fig:logits}.

From Figure \ref{fig:logits}, the distribution after being processed by SKD becomes smoother compared to the output of the teacher model. The value of the target class in the distribution is significantly lower than that of other classes. SKD makes the model output smoother and simpler for the student model, which makes it easier for the student model to learn. In addition, unlike the distillation temperature, SKD uses the learning simplifier to individually process each data point based on its similarity to other data in a batch. This allows SKD to obtain more finely tuned changes to the teacher logits compared to the distillation temperature, resulting in high distillation fidelity.
\paragraph{Comparison of distillation fidelity}
We use the average agreement between the predictions of the student model and the teacher model to measure the distillation fidelity \cite{dkdrw}. A higher average agreement reflects a more faithful imitation of the student model. The calculation of the average agreement for n data points is as follows:
\begin{equation}
\mathrm{Average\ Agreement}:=\frac{1}{n}\sum_{i=1}^{n}\mathbb{I}\{\mathop{\arg\max}\limits_{j}p_{i,j}^s=\mathop{\arg\max}\limits_{j}{p}_{i,j}^t\}.
\end{equation}
By comparing the distillation fidelity using SKD and KD, we can determine whether SKD makes it easier for student models to mimic the knowledge of teacher models. The comparison results are shown in Table \ref{tab:agree}.
\begin{table}[htbp]
\centering
\caption{Comparison results of the average agreement.\label{tab:agree}}
  \begin{tabular}{ccc}
    Method     & Training Set & Validation Set \\
    \midrule[1pt]
      KD & 0.86 & 0.75  \\
   	\textbf{SKD} & \textbf{0.92} & \textbf{0.79}      \\
  \end{tabular}
\end{table}

The experimental results are consistent with our design idea of SKD; that is, our SKD helps student models to mimic teacher models more easily, significantly improving the distillation fidelity and consequently enhancing the effect of knowledge distillation.

\begin{figure}[htbp]
  \subfigure[KD\label{fig:kdtsne}]
{
    \begin{minipage}{.47\linewidth}
        \centering
        \includegraphics[scale=0.5]{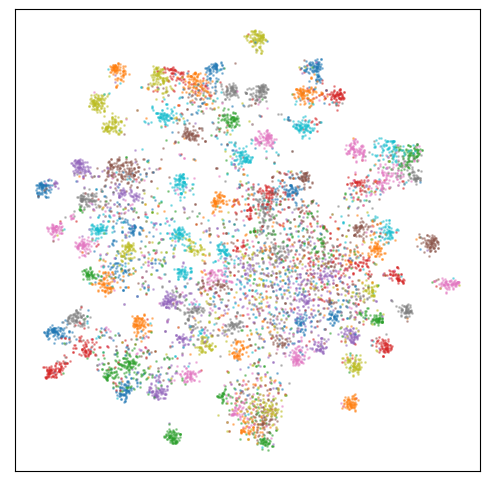}
    \end{minipage}
}
\subfigure[SKD\label{fig:skdtsne}]
{
 	\begin{minipage}{.49\linewidth}
        \centering
        \includegraphics[scale=0.5]{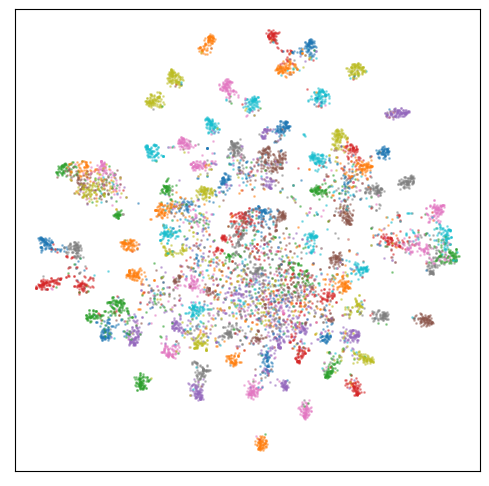}
    \end{minipage}
}
  \caption{t-SNE visualization of the student logits.\label{fig:tsne}}
\end{figure}

\paragraph{Features visualization}
We also visualized  the student model’s features using t-SNE \cite{tsne} as shown in Figure \ref{fig:tsne}. The features of the student model trained with SKD are more compact within the same category, and the differences between different categories are more pronounced. This proves that SKD enables the student model to learn clearer relationships between categories and perform more accurate classification.

\section{Conclusion}
In this paper, to improve the learning effect of a student model, we propose student-friendly knowledge distillation (SKD) to simplify the teacher output into new knowledge representations. Unlike other knowledge distillation methods, our SKD focuses on changing the knowledge output of the teacher model. First, we use softening processing to soften the output of the teacher model, making it easier for the student to learn. Then, through the learning simplifier, based on the mutual relationships of various data in the logits of the teacher model output, we simplify the logits to produce a new learning objective that is more suitable for the student model. This helps the student model mimic the teacher model's logits more easily, thereby enhancing the effect of distillation while maintaining high training efficiency. At the same time, SKD can be combined with other knowledge distillation methods, including logit-based and feature-based methods, to enhance the distillation effect. We hope this paper can inspire new research ideas regarding knowledge distillation.

\paragraph{Limitations}
SKD, as a logit-based knowledge distillation method, could not outperform state-of-the-art feature-based methods on object detection tasks due to the lack of location knowledge in the logits. Besides, the relationships between different combinations of teacher-student models and the best value of the parameter $\alpha$ cannot be determined in SKD currently. We plan to find a method to determine the optimal $\alpha$ in future work.
\bibliography{Ref1}
\bibliographystyle{ieee_fullname}

\newpage
\appendix
\section{Appendix}
\subsection{Experimental settings}
\label{app:seting}
Our experiments included various commonly used models, such as ShuffleNet \cite{shu2, shuffnet}, MobileNet-V2 \cite{mov2}, VGG \cite{verydeep}, ResNet \cite{res}, and WRN \cite{wrn}, as teacher and student models. Our experiments also compared feature-based methods, such as FitNet \cite{fitnet}, AT \cite{at}, RKD \cite{rkd}, CRD \cite{crd}, OFD \cite{ofd}, and ReviewKD \cite{reviewkd}, as well as logit-based methods, such as KD \cite{kd} and DKD \cite{dkd}. These methods are representative knowledge distillation methods and include the state-of-the-art feature-based and logit-based methods, namely, ReviewKD \cite{reviewkd} and DKD \cite{dkd}, respectively.

We compared our method with other state-of-the-art models on standard datasets, including the following:

\textbf{CIFAR-100} \cite{cifar} is a commonly used image classification dataset consisting of 60,000 32x32 images across 100 categories, with 50,000 images in the training set and 10,000 in the test set.

\textbf{ImageNet} \cite{imagenet} is a large-scale image classification dataset with 1,000 categories. The dataset includes over 1.2 million training images and 50,000 test images.

The details of the experiments were kept the same as in DKD [20]. The specific training details are as follows:

For \textbf{CIFAR-100} \cite{cifar}, we trained the models using the SGD optimizer for 240 epochs. The batch size used was 64. For ShuffleNet \cite{shu2, shuffnet} and MobileNet-V2 \cite{mov2}, we used an initial learning rate of 0.01, while for VGG \cite{verydeep}, ResNet \cite{res}, and WRN \cite{wrn} models, we used an initial learning rate of 0.05. Then, during the training process, the learning rate decayed by a factor of 10 at the 150th, 180th, and 210th epochs. The weight decay and momentum of the optimizer were set to 5e-4 and 0.9, respectively. The weight of the cross-entropy loss in the distillation loss was set to 1.0, and the temperature was set to 4. The linear warm-up in the training process was set to 20 epochs. For our learning simplifier, we set the dimensions of $q$,$k$ and $v$ in the self-attention to 512. The learning rate of the parameters in SKD was set to 3e-5, and the weight decay of the optimizer (SGD) was set to 5e-4. The dropout ratio in the output layer was set to 0.5. The value of the alpha coefficient in the distillation loss was adjusted for different teacher-student combinations.

For \textbf{ImageNet} \cite{imagenet}, we trained the models using the SGD optimizer for a total of 100 epochs. The batch size was 512. The initial learning rate was set to 0.2 and divided by 10 for every 30 epochs. The weight decay was set to 1e-4. The weight of the cross-entropy loss in the distillation loss was set to 1.0, and the temperature was set to 1. For the ImageNet dataset, the dimensions of $q$,$k$ and $v$ in the self-attention of SKD were set to 2048, while the other settings were the same as those for the CIFAR-100 dataset.

Our experiments on CIFAR-100 are trained with 1 NVIDIA V100, and the experiments on ImageNet are trained with 8 NVIDIA V100. 
\subsection{Related work}
\paragraph{Knowledge distillation}
The concept of knowledge distillation for deep learning models was proposed by Hinton et al. \cite{kd}. It allows a small student model to be trained simultaneously with the ground-truth labels and soft labels generated by the teacher model, which is softened using the distillation temperature. Most knowledge distillation methods can be classified as logit-based methods \cite{nkd, dkd, dist, dml, resoft, bann, takd, snapshot, oneff, dgkd} or feature-based methods \cite{ofd, crd, reviewkd, fitnet, rkd, at, ktab, pcn, sp, mgd, agift, cckd}.

Logit-based methods optimize the distillation loss function of the logits to enable the student model to learn from the teacher model more effectively. Among them, WSL \cite{resoft} weights the original distillation loss based on the relationships between the teacher and student logits to balance the bias-variance tradeoff during training. DKD \cite{dkd}, NKD \cite{nkd}, and DIST \cite{dist} analyze and optimize the original loss function, proposing new loss functions to flexibly control the teacher knowledge the student model needs to learn. TAKD \cite{takd} and DGKD \cite{dgkd} use an intermediate-sized "teacher assistant" model to help transfer the knowledge of the teacher model to the student model, thereby avoiding poor learning performance of the student model caused by large differences in the capacity between the teacher and student models.

On the other hand, feature-based methods enable the student to learn the features of the intermediate layers of the teacher model, thereby further boosting distillation performance. FitNet \cite{fitnet}, OFD \cite{ofd}, and other methods \cite{at, pcn, ktab} align the features of the student and teacher intermediate layers through the design of different conversion modules to achieve better knowledge transfer. CRD \cite{crd}, RKD \cite{rkd}, and other methods \cite{sp, agift} allow the student to learn the correlations among the teacher model's intermediate layer features to learn different aspects of the knowledge. In addition, Chen et al. \cite{reviewkd} used multilevel distillation to learn multilevel knowledge and improve the effectiveness of distillation. While feature-based methods generally achieve better results, they require a more complex design and greater computational and storage costs than logit-based methods.

In addition, some research focuses on the principles of knowledge distillation \cite{oneff, dkdrw, isls, rels, relskd}. Cho and Hariharan \cite{oneff} believe that due to the large discrepancy in capacity between the student and teacher models, student models cannot imitate teacher models well. Stanton et al. \cite{dkdrw} conducted in-depth research on the fidelity of student models imitating teacher models and summarized the reasons for low fidelity.

This paper focuses on logit-based methods. Previous methods focused on improving the distillation loss function or optimizing the distillation process, while our method focuses on the logits of the teacher model. By modifying the logits based on the correlations between the data and the student output, we aim to reduce the difficulty of learning of the student model to improve the distillation performance.
\paragraph{Attention}
The attention mechanism was proposed in sequence models to capture the correlations of sequences without considering distance \cite{att}. Self-attention computes internal attention within a sequence to obtain its representation \cite{seatt1, seatt2}. Many attention-based models have since been applied in various fields, achieving excellent results \cite{vit, trans, tdit, deepvit}.

Some feature-based distillation methods incorporate attention into their approach \cite{afd, cd, alp}. Among them, CD \cite{cd} uses the attention mechanism to weight each channel of the intermediate-layer features of the teacher, thereby highlighting the information in the more important channels for the student. AFD \cite{afd} uses attention to select which intermediate layer features of the teacher model need to be distilled to the student model. However, attention-based methods have yet to be used in logit-based distillation methods.

\subsection{More Analysis}
\label{app:ana}
\begin{figure}
    \centering
    \includegraphics[width=0.6\linewidth]{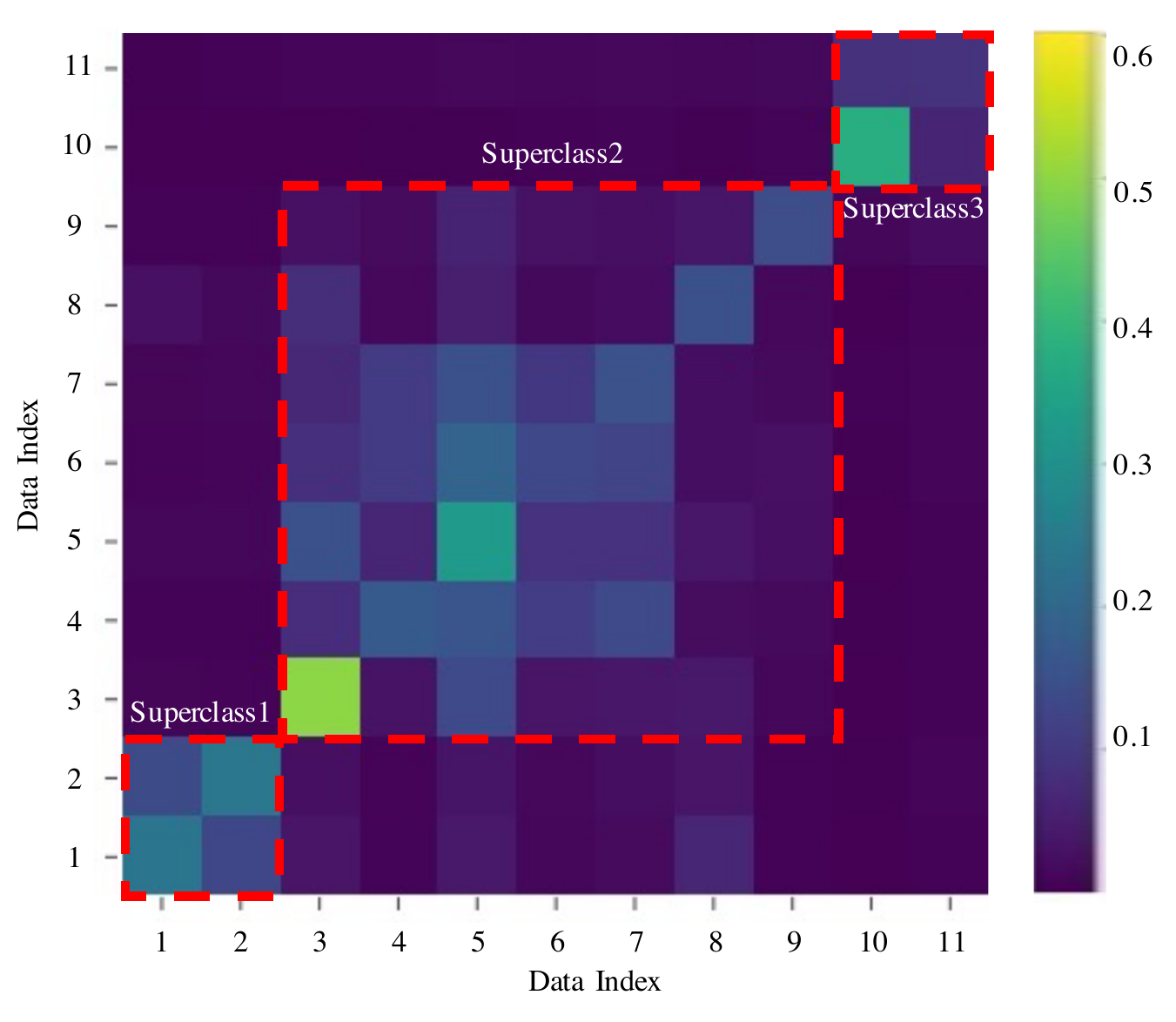}
  	\caption{Visualization of the heatmap of attention matrix A.\label{fig:attm}}
\end{figure}

\paragraph{Attention matrix analysis}
The attention matrix is crucial to the attention mechanism. By visualizing the attention matrix in the learning simplifier, we can see how attention weights are distributed in a batch of data. In the CIFAR-100 dataset, there are a total of 20 superclasses, each of which contains five classes. Therefore, for a given class, there are four classes that are similar to it (i.e., members of the same superclass) and 95 classes that are different. We sorted the attention matrix by superclass, randomly selected three superclasses from a batch of data, and created an intuitive visualization of the attention matrix, as shown in Figure \ref{fig:attm}.

In the heatmap, the attention weights for data of similar classes are larger than those of other classes. This indicates that for similar data, the original output distribution of the teacher model is similar. When using SKD to modify the logits, more attention is given to data of similar classes to obtain the change in the logits of the teacher model, which allows SKD to process each data distribution separately.

\paragraph{Efficiency analysis}

\begin{figure}
    \centering
    \includegraphics[width=0.6\linewidth]{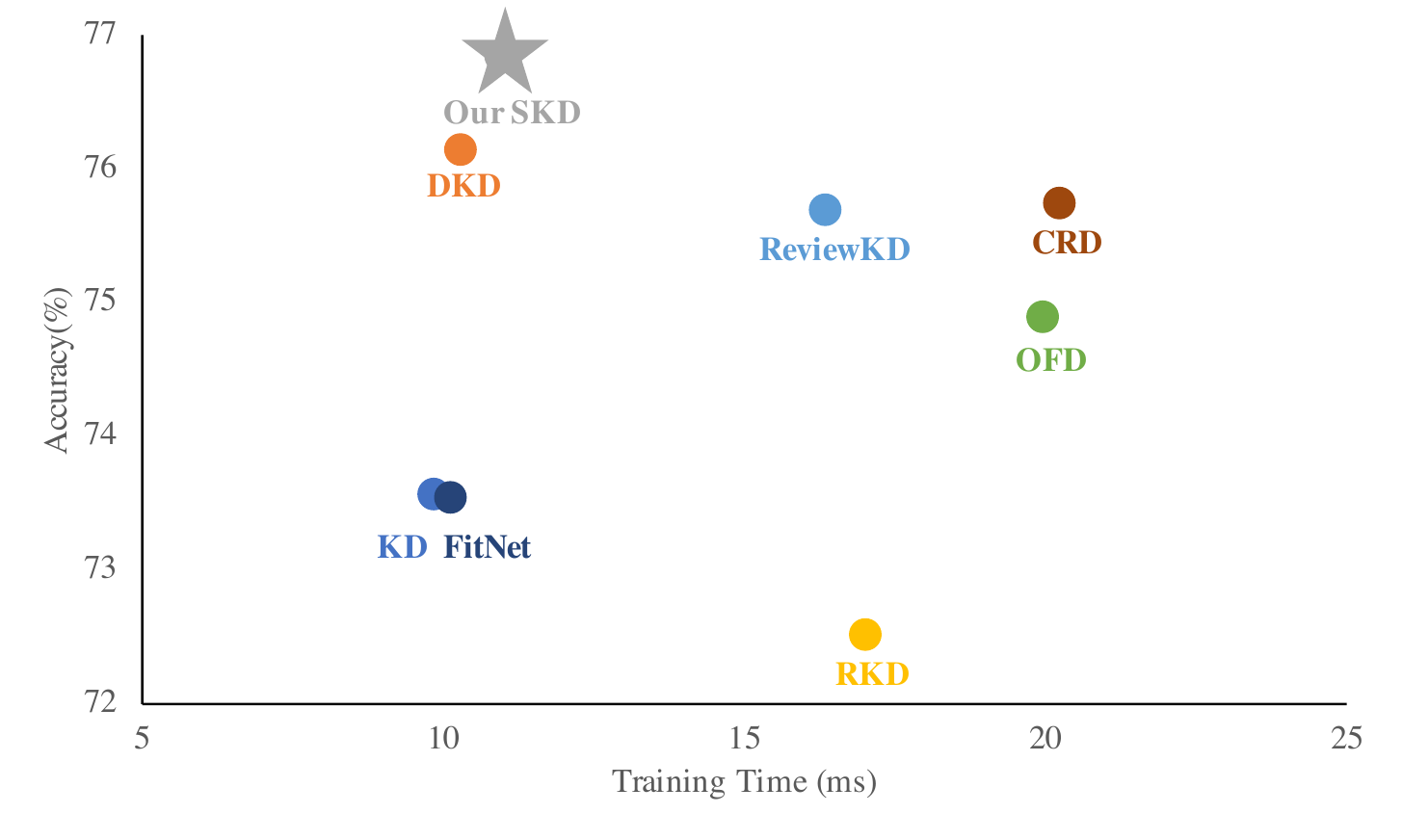}
  	\caption{Training efficiency analysis graph. The horizontal axis represents the training time for one batch of data, and the vertical axis represents the top-1 accuracy of the method.\label{fig:eff}}
\end{figure}

We analyzed the training efficiency of representative knowledge distillation methods to demonstrate the high efficiency of SKD. Since SKD targets the logits of the teacher model, it only requires simple processing after the teacher model generates the logit distribution, without the need for a complex knowledge transformation in the intermediate layer between the teacher and student models that feature-based methods employ. Therefore, the training efficiency of SKD is comparable to that of logit-based methods. As shown in Figure \ref{fig:eff}, our SKD achieves the best training effect, while the training time is close to that of the fastest method, KD.

\subsection{Broader Impacts}
Knowledge distillation  is a basic deep neural network training method, which prevents us from giving specific application impacts. However, since the student model needs to learn from the teacher model in knowledge distillation, we need to additionally consider the model security of the teacher model when we consider the model security of the student model, in addition to common factors, such as data security issues. Because a teacher model with security flaws is likely to train a student model with similar security issues. Therefore, we need to check the security of the teacher model before performing knowledge distillation to avoid being attacked because of the teacher model.

\end{document}